%% file: main.tex
\pdfoutput=1

\documentclass[11pt]{article}

\usepackage[]{EMNLP2022}

\usepackage{times}
\usepackage{latexsym}

\usepackage[T1]{fontenc}

\usepackage[utf8]{inputenc}

\usepackage{microtype}

\usepackage{inconsolata}

\usepackage{graphicx}
\usepackage{url}
\usepackage{latexsym}
\usepackage{makecell}
\usepackage{epstopdf}
\usepackage{tipa}
\usepackage{hyperref}
\usepackage{xstring}
\usepackage{amsfonts}
\usepackage{amsmath}
\usepackage{fixltx2e}
\usepackage[T1]{fontenc}
\usepackage{booktabs}
\usepackage{arabtex}
\usepackage{hhline}
\usepackage{pdfpages}
\usepackage{utf8}
\input{macros.tex}
\usepackage{multirow}
\usepackage{subfigure}
\usepackage{amssymb}
\usepackage[super]{nth}
\usepackage[export]{adjustbox}
\usepackage{wasysym}
\usepackage{footmisc}
\usepackage[super]{nth}
\newcommand{\APGC}{{\sc APGC}}

\setcode{utf8}
\vocalize
\definecolor{raspberry}{HTML}{eb0170}
\definecolor{purple(x11)}{HTML}{9c00fe}

\title{The Shared Task on Gender Rewriting}

\author{
    Bashar Alhafni,$^{1}$
    Nizar Habash,$^{1}$
    Houda Bouamor,$^{2}$
    Ossama Obeid,$^{1}$
    \\
    \textbf{Sultan Alrowili},$^{3}$ 
    \textbf{Daliyah Alzeer},$^{4}$
    \textbf{Khawlah M. Alshanqiti},$^{5}$
    \textbf{Ahmed ElBakry},$^{6}$\\
    \textbf{Muhammad ElNokrashy},$^{6}$
    \textbf{Mohamed Gabr},$^{6}$
    \textbf{Abderrahmane Issam},$^{7}$ \\
    \textbf{Abdelrahim Qaddoumi},$^{8}$
    \textbf{K. Vijay-Shanker},$^{3}$ 
    \textbf{Mahmoud Zyate}$^{9}$\Thanks{The first four authors are the shared task organizers, listed in order of contribution. The remaining authors are the shared task participants in alphabetical order.} \\
    $^{1}$New York University Abu Dhabi,
    $^{2}$Carnegie Mellon University in Qatar,\\
    $^{3}$University of Delaware,
    $^{4}$Taif University,
    $^{5}$Umm Alqura University,\\
    $^{6}$Microsoft ATL Cairo,
    $^{7}$Archipel Cognitive,
    $^{8}$New York University,
    $^{9}$Leyton \\
    \texttt{alhafni@nyu.edu} \\
}

\begin{document}
\maketitle
\begin{abstract}
In this paper, we present the results and findings of the Shared Task on Gender Rewriting, 
which was organized as part of the Seventh Arabic Natural Language Processing Workshop. 
The task of gender rewriting refers to generating alternatives of a given sentence to match different target user gender contexts (e.g., female speaker with a male listener, a male speaker with a male listener, etc.). This requires changing the grammatical gender (masculine or feminine) of certain words referring to the users. In this task, we focus on Arabic, a gender-marking morphologically rich language. 
A total of five teams from four countries participated in the shared task.
\end{list} 
\end{abstract}

\section{Introduction}
The problem of gender bias in Natural Language Processing (NLP) systems has been receiving a lot of attention across a variety of tasks such as machine translation, co-reference resolution, and dialogue systems. Research has shown that NLP systems do not only have the ability to embed societal biases, but they also amplify and propagate them in ways that create representational harms and degrade users' experiences \cite{sun-etal-2019-mitigating,blodgett-etal-2020-language}. The main cause of this problem is usually attributed to inherently biased data that is used to build these systems and which mirrors the inequalities of the world we live in. Therefore, many approaches were proposed to mitigate this problem by either using counterfactual data augmentation techniques \cite{lu2018gender,hall-maudslay-etal-2019-name,zmigrod-etal-2019-counterfactual} or by debiasing pretrained representation that is trained on biased data \cite{bolukbasi2016man,zhao-etal-2018-learning,manzini-etal-2019-black,zhao2020gender}. However, even the most balanced of models can still exhibit and amplify bias if they are designed to produce a single text output without taking their users’ gender preferences into consideration \cite{habash-etal-2019-automatic,alhafni-etal-2020-gender,alhafni-etal-2022-user}.  Therefore, to provide the correct user-aware output, NLP systems should be designed to produce outputs that are as gender specific as the users preferences they have access to. Recently, \newcite{alhafni-etal-2022-user} introduced the task of gender rewriting, which refers to generating alternatives of a given sentence to match different target user gender contexts. To encourage more researchers to work on this problem, we organized the Shared Task on Gender Rewriting. We focus on Modern Standard Arabic (MSA), a gender-marking morphologically rich language, in contexts involving two users.\footnote{\url{http://gender-rewriting-shared-task.camel-lab.com/}}

\begin{table*}[t]
\centering 
\includegraphics[width=0.9\linewidth]{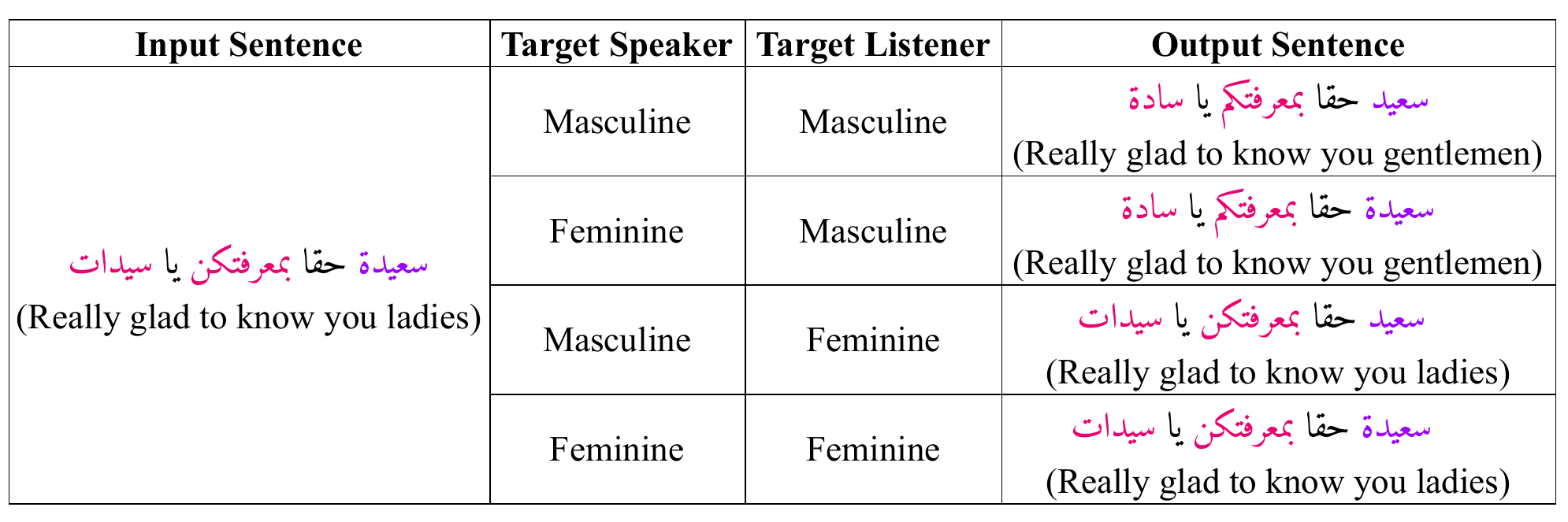}
\caption{Example of the gender rewriting task. The input sentence has four rewritten alternatives that match the different  target user gender contexts. First person gendered words are in \textcolor{purple(x11)}{purple} and second person gendered words are in \textcolor{raspberry}{red}.}
\label{tab:data-example}
\end{table*}

This shared task was organized as part of the Seventh Arabic Natural Language Processing Workshop (WANLP), collocated with EMNLP 2022. This is the first shared task at WANLP in seven years to target a language generation problem in Arabic. A total of five teams from four countries participated in the shared task. One team contributed to a system description paper which is included in the WANLP proceedings and cited in this paper. We provide a description of all submitted systems and the approaches they use. All of the datasets created for this shared task will be made publicly available to support further research on gender rewriting.

This paper is organized as follows. We first provide a description of the shared task (\S\ref{sec:task-desc}). We then describe the data used in the shared task, including a newly created set which we used for evaluation in \S\ref{sec:shared-task-data}. Next, we provide a description of all submitted systems in \S\ref{sec:systems} and discuss the results in \S\ref{sec:results}. Finally, we discuss the lessons we learned from running this shared task and provide recommendations to the (Arabic) NLP community in \S\ref{sec:lessons-learned}.



\section{Task Description}
\label{sec:task-desc}
The task of gender rewriting was introduced by \newcite{alhafni-etal-2022-user} and it refers to generating alternatives of a given Arabic sentence to match different target user gender contexts.  We focus on contexts involving two users (I and/or You) -- first and second grammatical persons with independent grammatical gender preferences. This requires changing the grammatical gender (masculine or feminine) of certain words referring to the users (speaker/first person and listener/second person) in the input sentence. Therefore, given an Arabic sentence as an input, the goal is to generate four different gender rewritten alternatives to match the different target user gender contexts (i.e., female speaker with a male listener, a male speaker with a male listener, a male speaker with a female listener, and a female speaker with a female listener). Table~\ref{tab:data-example} shows an example of the gender rewriting problem where the input sentence is rewritten to its four gender alternatives that match the four target user gender contexts.

\paragraph{Notation} We use the notation that is defined by \newcite{alhafni-etal-2022-user}. Namely, we use four elementary symbols to facilitate the discussion of this task: 1M, 1F, 2M and 2F. The digit part of the symbol refers to the grammatical person (\nth{1} or \nth{2}) and the letter part refers to the grammatical gender (Masculine or Feminine). Additionally, we  use B to refer to invariant/ambiguous gender.




\subsection{Shared Task Restrictions}
We provided the participants with a set of restrictions for building their systems to ensure a common experimental setup and fair comparison. Participants were asked
not to use any external manually labeled datasets. However, the use of publicly available unlabeled data was allowed. Participants were also not allowed to use the publicly available development and test sets of the shared task corpus for training their systems. Moreover, we provided the participants with a new blind test set that was manually annotated for this shared task. The participants were provided with the input sentences and they did not have access to the gold references. We discuss the properties and statistics of this new test set in more detail in \S\ref{sec:test-set}.

\subsection{Evaluation Metrics}
We follow \newcite{alhafni-etal-2022-user} by treating the gender rewriting problem as a user-aware grammatical error correction task and use the MaxMatch (M\textsuperscript{2}) scorer \cite{dahlmeier-ng-2012-better} as our evaluation metric. The M\textsuperscript{2} scorer computes the Precision (P), Recall (R), and F\textsubscript{0.5} by maximally matching phrase-level edits made by a system to gold-standard edits. The gold edits are computed by the M\textsuperscript{2} scorer based on provided gold references. We also report BLEU \cite{Papineni:2002:bleu} scores which are obtained using SacreBLEU \cite{post-2018-call}. We report the gender rewriting results in a normalized space for Alif, Ya, and Ta-Marbuta \cite{Habash:2010:introduction}.

\section{Shared Task Data}
\label{sec:shared-task-data}
In this section, we describe the data we use in the shared task.

\subsection{The Arabic Parallel Gender Corpus}
\label{sec:apgc}
We use the publicly available Arabic Parallel Gender Corpus ({\APGC}) -- a parallel corpus of Arabic sentences with gender annotations and gender rewritten alternatives of sentences selected from OpenSubtitles 2018 \cite{Lison:2016:opensubtitles2016}.
The corpus comes in three versions:  {\APGC} v1.0 \cite{habash-etal-2019-automatic}, {\APGC} v2.0 \cite{alhafni-etal-2022-arabic}, and {\APGC} v2.1 \cite{alhafni-etal-2022-user}. In this shared task, we use {\APGC} v2.1 which contains 80,326 gender-annotated parallel sentences (596,799 words) of contexts involving first and second grammatical persons covering singular, dual, and plural constructions.

\paragraph{Annotations} Each sentence in {\APGC} v2.1 has one of nine labels: 1M/2M, 1M/2F, 1F/2M, 1F/2F, 1M/B, B/2M, 1F/B, B/2F, and B. Each of these labels indicates the existence (or lack thereof) of first and/or second persons gendered references in the sentence. {\APGC} v2.1 also contains two types of word-level gender labels: basic and extended. The basic schema labels each word as B, 1F, 2F, 1M, or 2M. The basic labels refer to the \textit{primary} person-gender marking signal in the word, which could come from the base form if gendered or the pronominal enclitic if the base form is not gendered.\footnote{Changing the grammatical gender of Arabic words involves either changing the form of the base word, changing the pronominal enclitics that are attached to the base word, or a combination of both \cite{alhafni-etal-2022-user}} 
The extended schema  marks the person-genders of both the base words and their pronominal enclitics.
This results in 25 word-level gender labels (e.g., B+1F, 1F+2M, etc.). 
All sentences containing gender-specific words have gender-rewritten parallels. 
The parallels of B-labeled sentences are trivial copies.
Out of the 80,326 sentences in  {\APGC} v2.1, 
 54\% (43,346) contain gendered words.
In terms of word-level statistics, only 9.7\% (58,066) are gender specific.

{\APGC} v2.1 is organized into five parallel corpora that are fully aligned (1-to-1) at the word level: Input, Target 1M/2M, Target 1F/2M, Target 1M/2F, and Target 1F/2F. All five corpora are balanced in terms of gender, i.e., the number of 1F and 1M words is the same; and the number of 2F and 2M words is the same. The Input corpus contains sentences with all possible word types (B, 1F, 2F, 1M, 2M). The Target 1M/2M corpus contains sentences that consist of B, 1M, 2M words; the Target 1F/2M corpus contains sentences that consist of B, 1F, 2M words; the Target 1M/2F corpus contains sentences that consist of B, 1M, 2F words; and the Target 1F/2F corpus contains sentences that consist of B, 1F, 2F words. 


\paragraph{Splits} We use \newcite{alhafni-etal-2022-arabic}'s splits: 57,603 sentences (427,523 words) for training (TRAIN), 6,647 sentences (49,257 words) for development (DEV), and 16,076 sentences (120,019 words) for testing (TEST).

\begin{table*}[t]
\centering 
\includegraphics[width=\linewidth]{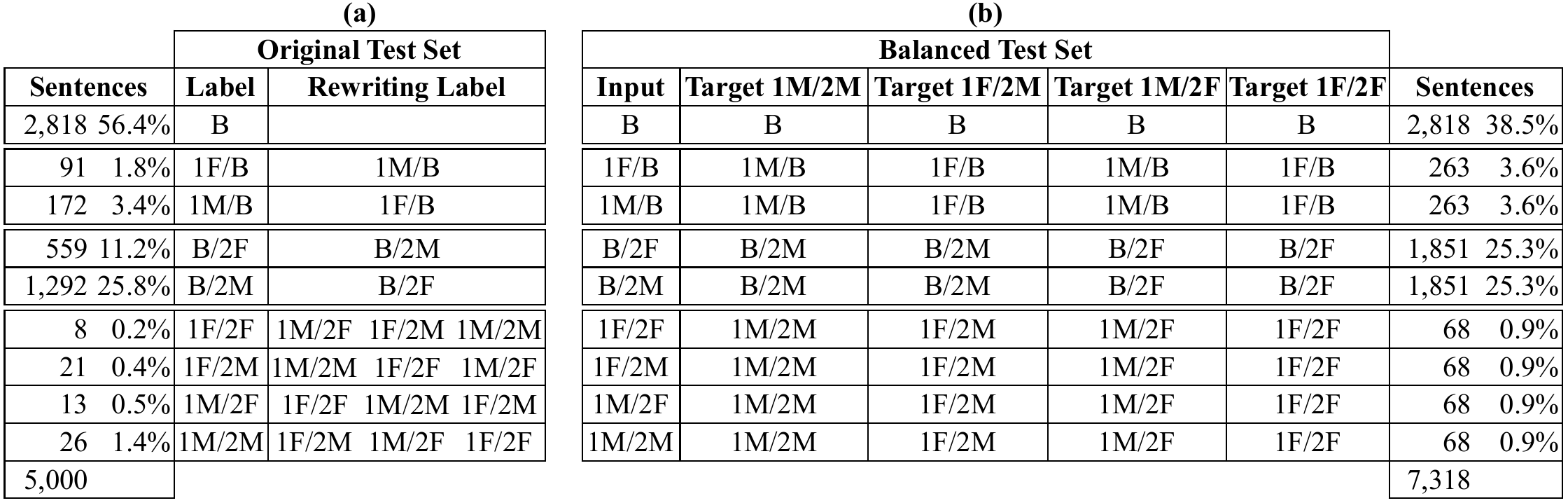}
\caption{\textbf{Sentence-level} statistics of the original (a) and the balanced Blind Test set (b) with its five versions.}
\label{tab:sent-level-stats}
\end{table*}

\begin{table*}[t]
\centering 
\includegraphics[width=\linewidth]{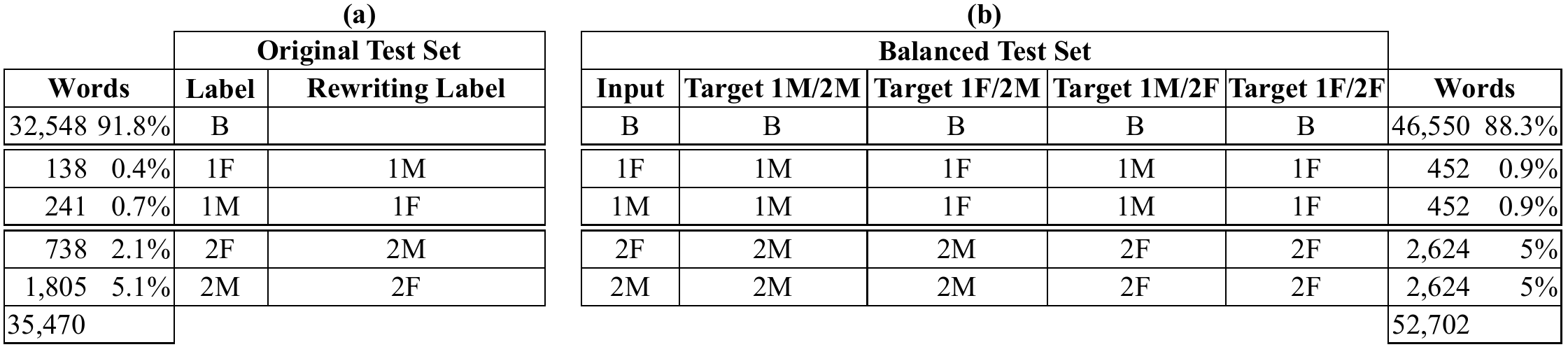}
\caption{\textbf{Word-level} statistics of the original (a) and the balanced Blind Test set (b) with its five versions.}
\label{tab:word-level-stats}
\end{table*}

\subsection{Blind Test Set}
\label{sec:test-set}
To ensure fair comparison between all participants, we manually annotated a new blind test set to evaluate their systems. We plan on making this new test set publicly available.
We will refer to this set as \textit{Blind Test} throughout the paper.

\paragraph{Data Selection} We followed the same procedure that was used in \cite{habash-etal-2019-automatic} and \cite{alhafni-etal-2022-arabic} to create the {\APGC}. We selected sentences from the English-Arabic OpenSubtitles 2018 dataset \cite{Lison:2016:opensubtitles2016} by extracting sentence pairs that include first or second pronouns on the English side. We annotated 5,000 sentences such that 1,061 (21.2\%) include first and second person pronouns, 2,116 (42.3\%) include only first person pronouns, and 1,823 (36.5\%) include only second person pronouns. The sentences were selected such: (a) they do not overlap with any of the sentences that are in {\APGC}; and (b) their proportions approximate the distribution of the Arabic-English pairs in the OpenSubtitles 2018 dataset that have first or second persons pronouns on the English side \cite{alhafni-etal-2022-arabic}.


\paragraph{Data Annotation} We conducted the annotation through a linguistic annotation firm that hired professional linguists to complete the task.\footnote{\url{https://www.ramitechs.com/}} We provided them with the same annotation guidelines that were defined in \newcite{alhafni-etal-2022-arabic} and used to annotate the {\APGC}. That is, the annotators were asked to identify the genders of the first and second person references in each sentence. In the case a gendered reference exists, the annotators were asked to copy the sentence and modify it to obtain the opposite gender forms. As was done when creating the {\APGC}, the modifications are strictly limited to morphological reinflections and word substitutions. Therefore, the total number of words is maintained along with a perfect alignment between each sentence and its parallel opposite gender forms. This allowed us to obtain basic and extended word-level gender annotations automatically as was done by \newcite{alhafni-etal-2022-arabic,alhafni-etal-2022-user}.


\paragraph{Data Statistics} Table~\ref{tab:sent-level-stats}(a) includes the statistics of the newly annotated sentences. This constitutes the Original Blind Test set. Out of all sentences in this set, 2,818 (56.4\%) are labeled as B. There are 1,851 sentences (37\%) that include only second-person gendered references (B/2F and B/2M). This is about five times more than sentences with only first-person gendered references (1F/B and 1M/B), which accounts for 5.3\% (263 sentences) of all sentences. Moreover, the number of sentences including first or second person masculine references is more than the ones including feminine references (1,292 B/2M vs 559 B/2F, and 172 1M/B vs 91 1F/B). There are 68 (1.4\%) sentences that have both first and second gendered references. These results are consistent with  {\APGC} v2.0  \cite{alhafni-etal-2022-arabic}.
The basic word-level statistics of the Original Blind Test set are presented in Table~\ref{tab:word-level-stats}(a). We evaluated inter-annotator agreement (IAA) on 500 sentences between two annotators. The IAA in terms of  nine sentence-level labels (B, M, F, for \nth{1} and for \nth{2} persons, e.g., 1M/2F or 1B/2M) was 98.0\%. Agreement in exact match on gender rewriting alternatives was 96.2\%. 

\input{extended_word_level}

Similarly to \newcite{habash-etal-2019-automatic} and \newcite{alhafni-etal-2022-arabic}, to ensure equal
gender representation in our dataset, we force balance the corpus by adding the manually rewritten sentences to the test set and  using their original forms as their rewritten forms. This constitutes the Balanced Blind Test set. The sentence-level statistics of the balanced set are presented in Table~\ref{tab:sent-level-stats}(b). This corpus has 7,318 sentences in total. Out of all sentences, 38.5\% (2,818) are marked as B, whereas sentences with gendered references constituted 61.5\% (4,500 sentences). Moreover, we organize the data into five balanced corpora as was done in {\APGC} v2.0 (\S\ref{sec:apgc}). The basic word-level statistics of the Balanced Blind Test set are presented in Table~\ref{tab:word-level-stats}(b). The extended word-level statistics of the Balanced Blind Test set are in Table~\ref{tab:word-level-extended-stats}.

\begin{table*}[t!]
\centering 
\setlength{\tabcolsep}{5pt}
\begin{tabular}{|l|l|}
\hline
\textbf{Team}           & \textbf{Affiliation} \\\hline \hline 
\textbf{Cairo Team}     & Microsoft ATL Cairo, Egypt \\\hline 
\textbf{CasaNLP}        & Archipel Cognitive, and  Leyton, Morocco \\\hline 
\textbf{Distinguishers} & Taif University, and Umm Alqura University, KSA \\\hline 
\textbf{Qaddoumi}       & New York University, USA \\\hline 
\textbf{UDEL-NLP}       & University of Delaware, USA \\\hline 
\end{tabular}

\caption{List of the five teams who participated in the gender rewriting shared task.}

\label{tab:teams}
\end{table*}

\begin{table*}[t!]
\centering 
 
\begin{tabular}{|l|l|l|l|}
\hline
\textbf{Team}           & \textbf{Gender ID} & \textbf{Special Preprocessing} & \textbf{Pretrained Models} \\\hline\hline
\textbf{Cairo Team}     & Word      & &  CAMeLBERT MSA + AraT5-MSA  \\\hline
\textbf{CasaNLP}        & Word      &  Word Side  Constraints &  CAMeLBERT MSA + AraT5-MSA \\\hline
\textbf{Distinguishers} & Word      & Morphological Features & CAMeLBERT MSA + AraBERT \\\hline
\textbf{Qaddoumi}       &           & Romanization & T5 \\\hline
\textbf{UDEL-NLP}       &           &  Sentence Side  Constraints  & ArabicT5\\\hline
\end{tabular}
\caption{Approaches and techniques used by the participants. Gender ID refers to gender identification. Special Preprocessing refers to any form of preprocessing done to modify the data (e.g., adding side-constraints, morphological processing, transliteration, etc.). Pretrained Models indicates the usage of pretrained models as part of the system.}
\label{tab:techniques-summary}
\end{table*}

\section{Participants and Systems}
\label{sec:systems}
Five teams from four countries participated in the shared task. Table~\ref{tab:teams} presents the names of the participating teams and their affiliations. Next, we describe the approaches the participants took to develop their gender rewriting systems.

\subsection{Systems Descriptions}
All participants leveraged pretrained language models such as
AraBERT~\cite{antoun-etal-2020-arabert}, CAMeLBERT~\cite{inoue-etal-2021-interplay}, T5~\cite{JMLR:v21:20-074}, and  AraT5~\cite{nagoudi-etal-2022-arat5},  when developing their systems. Some systems consisted of multiple components to do gender identification and then rewriting as was done in \newcite{alhafni-etal-2022-user}, while others treated the problem as a traditional sequence-to-sequence (Seq2Seq) task. Table~\ref{tab:techniques-summary} presents a summary of the different approaches used to develop the different systems.

\paragraph{Cairo Team} The system developed by Cairo~Team was a multi-step system  consisting of the following components: (a) a word-level gender identification classifier; (b) a word-level person identification classifier; and (c) sentence-level gender rewriting Seq2Seq models. The word-level classifiers were built by fine-tuning CAMeLBERT MSA~\cite{inoue-etal-2021-interplay}, on the training data of {\APGC} v2.1. Cairo~Team used the \textit{basic} word-level annotations in the corpus to build these two classifiers. Concretely, the gender identification component was trained to identify the gender of each word as M, F, or B, whereas the person identification component was trained to classify the person which the word refers to as \nth{1}, \nth{2}, or none. For the sentence-level Seq2Seq models, Cairo~Team  built four different models, one for each target user gender context (i.e., 1M/2M, 1F/2M, 1M/2F, 1F/2F), by fine-tuning AraT5-MSA\textsubscript{BASE}~\cite{nagoudi-etal-2022-arat5}.

During inference, the input sentence is passed to the word-level classifiers to get the gender and person labels for each word. These predicted labels indicate which words need to be rewritten based on the compatibility between the labels and the target user gender contexts.
Then, the same input sentence is passed to each Seq2Seq model to get its rewritten forms. After that, Cairo~Team  uses a simple heuristic to reduce the noise that could be generated in the outputs of the Seq2Seq models and to ensure that only the necessary gendered words are changed. To do so, Cairo~Team  generates all subsets of possible trigrams for each gendered word that needs to be changed in the input. Then, they search for partial matches of these trigrams in the Seq2Seq model generated sentences and pick the generated words that have the highest match. The intuition behind this approach is that: (a) the Seq2Seq model would benefit from seeing the entire sentence to apply in-context word gender rewriting; and  (b) most of the gendered words in the {\APGC} v2.1 (96.9\%) are due to morphological inflections, which allows the matching heuristic to have a high coverage.

The fine-tuning of the models was done using Hugging~Face's Transformers~\cite{wolf-etal-2020-transformers}. Both the word-level gender and person identification classifiers were fine-tuned on a single GPU for 10 epochs with a maximum sequence length of 128, a batch size of 32, and a learning rate of 1e-4. The sentence-level gender rewriting component was fine-tuned on a single GPU for 30 epochs with a maximum sequence length of 128, a batch size of 16, and a learning rate of 1e-3. Checkpoints were saved every 1000 steps and at the end of fine-tuning, the best checkpoint was picked based on the development set.

\paragraph{CasaNLP} The system introduced by CasaNLP was also a multi-step system that consists of word-level gender identification and sentence-level gender rewriting. For gender identification, the team used the gender identification model that was developed and released by \newcite{alhafni-etal-2022-user}.\footnote{\url{https://github.com/CAMeL-Lab/gender-rewriting/}\label{alhafni-code}} The gender identification component takes the input sentence and assigns an \textit{extended} gender label to every word in the input. After that and based on the compatibility between the labels and the target user gender contexts, CasaNLP  adds word-level target gender labels as \textit{side-constraints}~\cite{sennrich-etal-2016-controlling} to the words that need to be rewritten in the input sentence (e.g., \<سعيد>~\LR{[2F]}~\<أنا>). They do this preprocessing step across all sentences in {\APGC}~v2.1. Then, they fine-tune AraT5-MSA\textsubscript{BASE} on the preprocessed sentences in TRAIN. The intuition here is that the model should learn to only rewrite the words that are marked in the input.  The team follows the same procedure during inference to generate the gender rewritten alternatives.

The fine-tuning of the models was done using Hugging Face's Transformers. The sentence-level gender rewriting system was fine-tuned for 10 epochs with a maximum sequence length of 64, a batch size of 32, and a learning rate of 1e-3 with 4 gradient accumulation steps.

\paragraph{Distinguishers} This team introduced a multi-step system that does word-level gender identification and out-of-context word-level gender rewriting. For gender identification, they used the model that was developed and released by \newcite{alhafni-etal-2022-user}.\footref{alhafni-code} For gender rewriting, the team developed an out-of-context word-level Seq2Seq model. The model followed the approach introduced in BERT-fused~\cite{zhu-etal-2020-bert-fused}, where they first use AraBERT~\cite{antoun-etal-2020-arabert} to extract representations for the input word, and then the representations are fused with each layer of the encoder and decoder of a standard Transformer model \cite{Vaswani:2017:attention}. The model was trained on gendered words present in {\APGC} v2.1. They also explored adding morphological features to their Seq2Seq model. They used CAMeLTools~\cite{obeid-etal-2020-camel} to do morphological tokenization on the words and get their part-of-speech tags. They added the tags as side-constraints to each word. During inference, they first run the gender-identification component over the input sentence to get predicted gender labels for each word. Then for each word that needs to be rewritten, they pass it to the Seq2Seq model to get its gender alternative. 

The out-of-context word-level gender rewriting model was built using Simple Transformers.\footnote{\url{https://github.com/ThilinaRajapakse/simpletransformers}} The model was fine-tuned on a single GPU for 5 epochs with a maximum sequence length of 25, a learning rate of 1e-5, and a batch size of 32.

\paragraph{Qaddoumi} The approach this team took to build their gender rewriting system relied on romanizing the Arabic text and using an \textit{English} pretrained model. The team preprocessed the data in  {\APGC}~v2.1 by using the Safe Buckwalter transliteration scheme \cite{Buckwalter:2002:buckwalter,Habash:2010:introduction}. They continue fine-tuning a grammatical error correction model that was originally built by fine-tuning T5~\cite{JMLR:v21:20-074} on the JFLEG corpus \cite{napoles-etal-2017-jfleg}.\footnote{\url{https://huggingface.co/vennify/t5-base-grammar-correction}} When producing the final outputs, they convert the text back to Arabic script. 

The sentence-level gender rewriting system was fine-tuned using the Happy Transformer library on a single GPU for 5 epochs with a maximum sequence length of 1024, a batch size of 32, and a learning rate of 5e-5.\footnote{\url{https://github.com/EricFillion/happy-transformer}} 

\input{results_test}

\paragraph{UDEL-NLP} The system developed by UDEL-NLP was at the sentence-level and based on T5. The team introduced  a new Arabic T5 model called ArabicT5~\cite{alrowili-etal-2022-shared-task}, which was pretrained on MSA by using an efficient T5 implementation \cite{tay:2021:efficient}. They fine-tuned the ArabicT5 model by adding side-constraints to the beginning of each sentence to indicate the target users' gender, and appending an \texttt{<eos>} to each sentence. The team follows the same preprocessing steps during inference.

The sentence-level gender rewriting system was built by fine-tuning ArabicT5 using Hugging Face's Transformers on a single GPU for 70 epochs, a maximum sequence length of 512, a batch size of 32, and a learning of 1e-4.




\section{Results}
\label{sec:results}
Table~\ref{tab:test-set-results} presents the results on the newly annotated Blind Test set. The last row is for the state-of-the-art system by \newcite{alhafni-etal-2022-user}. The best result in terms of F\textsubscript{0.5} is achieved by the Cairo Team (75.42), the official winner of the shared task. This is mainly due to their high score in precision (76.26). Qaddoumi comes in second place achieving an F\textsubscript{0.5} of 59.68, followed by UDEL-NLP in third place with 59.08 in F\textsubscript{0.5}. In fourth place, CasaNLP achieves an F\textsubscript{0.5} score of 55.45 with the highest recall of 84.60. Distinguishers comes in fifth place, achieving 20.52 in F\textsubscript{0.5}. It is worth noting that none of the systems is able to beat the previously published system by \newcite{alhafni-etal-2022-user} applied to the new Blind Test.

\input{error_analysis}

\paragraph{Error Analysis} We conducted a simple error analysis over the outputs of all system on the Blind Test set. Given that most teams employed sentence-level Seq2Seq models when developing their gender rewriting systems, we suspected that the outputs will be noisy since sentence-level models will not guarantee that changes are only applied to gendered words, or maintain the word-level parallelism between the input and output. Table~\ref{tab:error-analysis}(a) presents the relative difference in the number of generated words for each team in comparison with the Blind Test reference; and Table~\ref{tab:error-analysis}(b) presents their correlation with the shared task metrics.
None of the teams maintained the total number of words. We observe a strong negative correlation between the absolute value of relative word count differences and the evaluation metrics -- almost -51\% correlation with  F\textsubscript{0.5}, and -78\% correlation with recall.


\input{results_test_no_pnx}

After inspecting the outputs of the submitted systems, we noticed that much of the noise was due to not handling punctuation correctly. We removed the punctuation from all the outputs and evaluated the systems in this space. Table~\ref{tab:test-set-results-no-pnx} shows the results on the Blind Test set after removing the punctuation. The scores of all teams went up significantly, with the exception of Distinguishers.  The highest increase of 31.6 points in F\textsubscript{0.5} is in the case of CasaNLP. In terms of the ranks of the systems in this unofficial evaluation space, CasaNLP is the best performer and they achieve 87.04 in F\textsubscript{0.5}. They also have the highest precision, recall, and BLEU scores. The Cairo Team comes in  second place with an F\textsubscript{0.5} of 83.76, followed by UDEL-NLP who achieves an F\textsubscript{0.5} of 70.22. Qaddoumi and Distinguishers are in fourth and fifth places, achieving 63.35 and 20.41 in F\textsubscript{0.5}, respectively.

\section{Outlook and Lessons Learned}
\label{sec:lessons-learned}
We organized this shared task on gender rewriting for Arabic  to raise  awareness in the Arabic NLP community of the problem of gender bias in Arabic NLP systems, and to encourage the community to come up with new approaches to alleviate this problem. Although the shared task received some interest from the community, the participation was limited\footnote{While 15 teams registered for the shared task  initially,  only five of them ended up participating.} when compared to other shared tasks organized at recent editions of WANLP\footnote{\url{http://www.arabic-nlp.net/}} or OSACT.\footnote{\url{https://osact-lrec.github.io/}} We believe that this is due to a couple of factors.

First is the \textbf{skewed interest towards sentence-level classification} tasks within the Arabic NLP community and the lack of novel open-vocabulary sequence transduction tasks. For instance,  most of the shared tasks organized at WANLP over the past few years focused on sentence-level classification to tackle  dialect identification: MADAR and NADI~\cite{bouamor-etal-2019-madar,abdul-mageed-etal-2020-nadi,abdul-mageed-etal-2021-nadi}; or Arabic sarcasm detection: ArSarcasm~\cite{abu-farha-etal-2021-overview}. The last shared task that featured a generation problem in Arabic was the QALB shared task on grammatical error correction \cite{rozovskaya-etal-2015-second}.

We acknowledge the importance of working on sentence-level classification problems,  but there are many natural language generation tasks where Arabic is still lagging behind compared to other languages. Examples of such tasks include dialectal machine translation, grammatical error correction, text simplification, and style transfer, to name a few. We envision that the development of resources and models for such tasks would re-spark the interest of the Arabic NLP community in a wide range of exciting, yet unsolved problems in Arabic NLP.

Second is the \textbf{novelty and difficulty} of the gender rewriting problem compared to other conventional sequence transduction tasks. Approaching the problem correctly requires developing controlled generation models that are able to make subtle, yet complex and grammatically correct, edits at the word level. 
%
In retrospect, we recognize that we could have organized this shared task  as two subtasks: one on gender identification at the word or sentence levels, and the other on sentence-level gender rewriting. This could have served as a bridge between classification and generation tasks, too, and allowed more people to participate for part if not the whole of the task.
As such,  we recommend that organizers of novel and nontraditional tasks to break the problem into subtasks to encourage more participation.

Lastly, the main goal of participating in a shared task is to learn about a new problem by introducing an interesting solution, which could benefit the community as a whole, as a positive or negative result. 
Being on top of the leaderboard should not be the only motive; we encourage organizers within the community to echo this sentiment when running their shared tasks.






\section*{Limitations and Ethical Considerations}
Our  intention of organizing this shared task is to increase the inclusiveness of NLP applications that deal with gender-marking morphologically rich languages. 
However, we acknowledge that, like all NLP technologies, developing systems for gender identification and rewriting could be used in malicious ways to discriminate against, or  erase, certain identities in certain contexts. 
We also acknowledge that by limiting the choice of gender expressions to  grammatical gender, we exclude  alternatives such as non-binary gender or no-gender expressions. We are not aware of any sociolinguistics published research that discusses such alternatives for Arabic. 
We stress on the importance of adapting Arabic NLP models to new gender alternative forms as they emerge as part of the language usage.


\bibliography{anthology,extra,camel-bib-v2}
\bibliographystyle{acl_natbib}

\end{document}

%% file: macros.tex
\newcommand{\hide}[1]{}



%% file: extended_word_level.tex
\tabcolsep=3.6pt
\begin{table}[t]
    \centering
    \footnotesize
    \begin{tabular}{|c|c|rl|}
\cline{1-2}
         \multicolumn{2}{|c|}{\textbf{Word Gender Label}} & \multicolumn{1}{c}{} \\\hline
         \textbf{Basic} & \textbf{Extended} & \multicolumn{2}{c|}{\textbf{Words}} \\\hline
        B & B & 46,550 & 88.3\% \\\hline\hline
        \multirow{2}{*}{1M} & 1M+B & 445 & 0.8\%\\
           & B+1M & 7 & 0.01\%\\\hline
        \multirow{2}{*}{1F} &  1F+B & 445 & 0.8\%\\
         & B+1F & 7 & 0.01\% \\\hline\hline
         \multirow{3}{*}{2M} & 2M+B & 2,464 & 4.7\%\\
             & B+2M & 144 & 0.3\%\\ 
         &  2M+2M & 16 & 0.03\%\\\hline

       \multirow{3}{*}{2F} &  2F+B & 2,464 & 4.7\%\\
    & B+2F & 144 & 0.3\%\\
       &  2F+2F & 16 & 0.03\% \\\hline
         \multicolumn{2}{r|}{}  &  \multicolumn{2}{l|}{52,702} \\\cline{3-4}
         
    \end{tabular}
    \caption{Statistics of the extended word-level gender of the Blind Test set.}
    \label{tab:word-level-extended-stats}

\end{table}

%% file: results_test.tex
\begin{table*}[ht!]
    \centering
    \begin{tabular}{|l|c|c|c|c|}
    \cline{1-5}
     \textbf{Team} & \textbf{Precision} & \textbf{Recall} & \textbf{F\textsubscript{0.5}} & \textbf{BLEU} \\\hline\hline
         \textbf{Cairo Team} & \textbf{76.26 (1)}& 72.27 (3) & \textbf{75.42 (1)}& \textbf{94.89 (1)}\\\hline
         \textbf{CasaNLP} & 51.05 (4) & \textbf{84.60 (1)} & 55.45 (4) & 86.06 (4) \\\hline
         \textbf{Distinguishers} & 20.93 (5) & 19.03 (5) & 20.52 (5) & 84.89 (5) \\\hline
         \textbf{Qaddoumi} & 56.49 (3) & 77.06 (2) & 59.68 (2) & 88.53 (3) \\\hline
         \textbf{UDEL-NLP} & 57.10 (2) & 68.61 (4) & 59.08 (3) & 91.02 (2)\\\hline\hline
         \textbf{\newcite{alhafni-etal-2022-user}} & \textbf{88.50} & \textbf{84.98} & \textbf{87.78} & \textbf{97.62} \\\hline

    \end{tabular}
    \caption{Results on the Blind Test set. Numbers in parentheses are the ranks.}
    \label{tab:test-set-results}
\end{table*}

%% file: error_analysis.tex
\begin{table}[]
    \setlength{\tabcolsep}{2.5pt}
    \centering
    \begin{tabular}{c | c}
    \begin{tabular}{|l|r|}
    \multicolumn{2}{c}{\textbf{(a)}}\\
     \hline
     \textbf{Team} &	\textbf{Word $\Delta$}\\\hline\hline
\textbf{Cairo Team} &	0.80\%\\\hline
\textbf{CasaNLP}	& -0.02\%\\\hline
\textbf{Distinguishers} &	1.28\%\\\hline
  \textbf{Qaddoumi} &	-0.63\%\\\hline
 \textbf{UDEL-NLP} &	0.05\%\\\hline
    \end{tabular}
         & 
\begin{tabular}{|l|r|}
    \multicolumn{2}{c}{\textbf{(b)}}\\
\hline
    \textbf{Metric} & \multicolumn{1}{c|}{\textbf{Correl}} \\\hline\hline
\textbf{Precision} & -42.95\%\\\hline
\textbf{Recall}  & 	-77.56\%\\\hline
\textbf{F\textsubscript{0.5}}  & 	-50.86\%\\\hline
\textbf{BLEU}  & 	-11.86\%\\\hline
\multicolumn{2}{c}{}\\
\end{tabular}

    \end{tabular}
    \caption{(a) The relative difference in the number of generated words for each team in comparison with the Blind Test reference. (b) The Pearson correlation of the shared task metrics in Table~\ref{tab:test-set-results} with the \textit{absolute} values of Word $\Delta$.  }
    \label{tab:error-analysis}
\end{table}

%% file: results_test_no_pnx.tex
\begin{table*}[ht!]

    \centering
    \begin{tabular}{|l|c|c|c|c|}
    \cline{1-5}
     \textbf{Team} & \textbf{Precision} & \textbf{Recall} & \textbf{F\textsubscript{0.5}} & \textbf{BLEU} \\\hline\hline
         \textbf{Cairo Team} & 87.34 (2)& 71.98 (3)& 83.76 (2)& 95.74 (2)\\\hline
         \textbf{CasaNLP} & \textbf{87.72 (1)} & \textbf{84.45 (1)} & \textbf{87.04 (1)}& \textbf{97.18 (1)}\\\hline
         \textbf{Distinguishers} & 20.81 (5) & 18.96 (5) & 20.41 (5) & 84.11 (5) \\\hline
         \textbf{Qaddoumi} & 60.68 (4)& 76.90 (2)& 63.35 (4) & 89.06 (4)\\\hline
         \textbf{UDEL-NLP} & 70.67 (3)& 68.50 (4)& 70.22 (3)& 91.99 (3)\\\hline\hline
         \textbf{\newcite{alhafni-etal-2022-user}} & \textbf{88.38} & \textbf{84.87} & \textbf{87.65} & \textbf{97.30} \\\hline
         
    \end{tabular}
    \caption{Results on the Blind Test set of after removing the punctuation. Numbers in parentheses are the ranks.}
    \label{tab:test-set-results-no-pnx}
\end{table*}

%% file: main.bbl
\begin{thebibliography}{36}
\expandafter\ifx\csname natexlab\endcsname\relax\def\natexlab#1{#1}\fi

\bibitem[{Abdul-Mageed et~al.(2020)Abdul-Mageed, Zhang, Bouamor, and
  Habash}]{abdul-mageed-etal-2020-nadi}
Muhammad Abdul-Mageed, Chiyu Zhang, Houda Bouamor, and Nizar Habash. 2020.
\newblock \href {https://aclanthology.org/2020.wanlp-1.9} {{NADI} 2020: The
  first nuanced {A}rabic dialect identification shared task}.
\newblock In \emph{Proceedings of the Fifth Arabic Natural Language Processing
  Workshop}, pages 97--110, Barcelona, Spain (Online). Association for
  Computational Linguistics.

\bibitem[{Abdul-Mageed et~al.(2021)Abdul-Mageed, Zhang, Elmadany, Bouamor, and
  Habash}]{abdul-mageed-etal-2021-nadi}
Muhammad Abdul-Mageed, Chiyu Zhang, AbdelRahim Elmadany, Houda Bouamor, and
  Nizar Habash. 2021.
\newblock \href {https://aclanthology.org/2021.wanlp-1.28} {{NADI} 2021: The
  second nuanced {A}rabic dialect identification shared task}.
\newblock In \emph{Proceedings of the Sixth Arabic Natural Language Processing
  Workshop}, pages 244--259, Kyiv, Ukraine (Virtual). Association for
  Computational Linguistics.

\bibitem[{Abu~Farha et~al.(2021)Abu~Farha, Zaghouani, and
  Magdy}]{abu-farha-etal-2021-overview}
Ibrahim Abu~Farha, Wajdi Zaghouani, and Walid Magdy. 2021.
\newblock \href {https://aclanthology.org/2021.wanlp-1.36} {Overview of the
  {WANLP} 2021 shared task on sarcasm and sentiment detection in {A}rabic}.
\newblock In \emph{Proceedings of the Sixth Arabic Natural Language Processing
  Workshop}, pages 296--305, Kyiv, Ukraine (Virtual). Association for
  Computational Linguistics.

\bibitem[{Alhafni et~al.(2020)Alhafni, Habash, and
  Bouamor}]{alhafni-etal-2020-gender}
Bashar Alhafni, Nizar Habash, and Houda Bouamor. 2020.
\newblock \href {https://www.aclweb.org/anthology/2020.gebnlp-1.12}
  {Gender-aware reinflection using linguistically enhanced neural models}.
\newblock In \emph{Proceedings of the Second Workshop on Gender Bias in Natural
  Language Processing}, pages 139--150, Barcelona, Spain (Online). Association
  for Computational Linguistics.

\bibitem[{Alhafni et~al.(2022{\natexlab{a}})Alhafni, Habash, and
  Bouamor}]{alhafni-etal-2022-arabic}
Bashar Alhafni, Nizar Habash, and Houda Bouamor. 2022{\natexlab{a}}.
\newblock \href {https://aclanthology.org/2022.lrec-1.199} {The {A}rabic
  parallel gender corpus 2.0: Extensions and analyses}.
\newblock In \emph{Proceedings of the Thirteenth Language Resources and
  Evaluation Conference}, pages 1870--1884, Marseille, France. European
  Language Resources Association.

\bibitem[{Alhafni et~al.(2022{\natexlab{b}})Alhafni, Habash, and
  Bouamor}]{alhafni-etal-2022-user}
Bashar Alhafni, Nizar Habash, and Houda Bouamor. 2022{\natexlab{b}}.
\newblock \href {https://aclanthology.org/2022.naacl-main.46} {User-centric
  gender rewriting}.
\newblock In \emph{Proceedings of the 2022 Conference of the North American
  Chapter of the Association for Computational Linguistics: Human Language
  Technologies}, pages 618--631, Seattle, United States. Association for
  Computational Linguistics.

\bibitem[{Alrowili and Vijay-Shanker(2022)}]{alrowili-etal-2022-shared-task}
Sultan Alrowili and K.~Vijay-Shanker. 2022.
\newblock Generative approach for gender-rewriting task with {A}rabic{T}5.
\newblock In \emph{Proceedings of the Seventh Arabic Natural Language
  Processing Workshop}, Abu Dhabi, UAE. Association for Computational
  Linguistics.

\bibitem[{Antoun et~al.(2020)Antoun, Baly, and Hajj}]{antoun-etal-2020-arabert}
Wissam Antoun, Fady Baly, and Hazem Hajj. 2020.
\newblock \href {https://www.aclweb.org/anthology/2020.osact-1.2} {{A}ra{BERT}:
  Transformer-based model for {A}rabic language understanding}.
\newblock In \emph{Proceedings of the 4th Workshop on Open-Source Arabic
  Corpora and Processing Tools, with a Shared Task on Offensive Language
  Detection}, pages 9--15, Marseille, France. European Language Resource
  Association.

\bibitem[{Blodgett et~al.(2020)Blodgett, Barocas, Daum{\'e}~III, and
  Wallach}]{blodgett-etal-2020-language}
Su~Lin Blodgett, Solon Barocas, Hal Daum{\'e}~III, and Hanna Wallach. 2020.
\newblock \href {https://doi.org/10.18653/v1/2020.acl-main.485} {Language
  (technology) is power: A critical survey of {``}bias{''} in {NLP}}.
\newblock In \emph{Proceedings of the 58th Annual Meeting of the Association
  for Computational Linguistics}, pages 5454--5476, Online. Association for
  Computational Linguistics.

\bibitem[{Bolukbasi et~al.(2016)Bolukbasi, Chang, Zou, Saligrama, and
  Kalai}]{bolukbasi2016man}
Tolga Bolukbasi, Kai-Wei Chang, James~Y Zou, Venkatesh Saligrama, and Adam~T
  Kalai. 2016.
\newblock \href
  {https://proceedings.neurips.cc/paper/2016/file/a486cd07e4ac3d270571622f4f316ec5-Paper.pdf}
  {Man is to computer programmer as woman is to homemaker? debiasing word
  embeddings}.
\newblock In \emph{Advances in Neural Information Processing Systems},
  volume~29. Curran Associates, Inc.

\bibitem[{Bouamor et~al.(2019)Bouamor, Hassan, and
  Habash}]{bouamor-etal-2019-madar}
Houda Bouamor, Sabit Hassan, and Nizar Habash. 2019.
\newblock \href {https://doi.org/10.18653/v1/W19-4622} {The {MADAR} shared task
  on {A}rabic fine-grained dialect identification}.
\newblock In \emph{Proceedings of the Fourth Arabic Natural Language Processing
  Workshop}, pages 199--207, Florence, Italy. Association for Computational
  Linguistics.

\bibitem[{Buckwalter(2002)}]{Buckwalter:2002:buckwalter}
Tim Buckwalter. 2002.
\newblock Buckwalter {{A}rabic} morphological analyzer version 1.0.
\newblock Linguistic Data Consortium (LDC) catalog number LDC2002L49, ISBN
  1-58563-257-0.

\bibitem[{Dahlmeier and Ng(2012)}]{dahlmeier-ng-2012-better}
Daniel Dahlmeier and Hwee~Tou Ng. 2012.
\newblock \href {https://www.aclweb.org/anthology/N12-1067} {Better evaluation
  for grammatical error correction}.
\newblock In \emph{Proceedings of the 2012 Conference of the North {A}merican
  Chapter of the Association for Computational Linguistics: Human Language
  Technologies}, pages 568--572, Montr{\'e}al, Canada.

\bibitem[{Habash et~al.(2019)Habash, Bouamor, and
  Chung}]{habash-etal-2019-automatic}
Nizar Habash, Houda Bouamor, and Christine Chung. 2019.
\newblock \href {https://doi.org/10.18653/v1/W19-3822} {Automatic gender
  identification and reinflection in {A}rabic}.
\newblock In \emph{Proceedings of the First Workshop on Gender Bias in Natural
  Language Processing}, pages 155--165, Florence, Italy.

\bibitem[{Habash(2010)}]{Habash:2010:introduction}
Nizar~Y Habash. 2010.
\newblock \emph{Introduction to {A}rabic natural language processing},
  volume~3.
\newblock Morgan \& Claypool Publishers.

\bibitem[{Hall~Maudslay et~al.(2019)Hall~Maudslay, Gonen, Cotterell, and
  Teufel}]{hall-maudslay-etal-2019-name}
Rowan Hall~Maudslay, Hila Gonen, Ryan Cotterell, and Simone Teufel. 2019.
\newblock \href {https://doi.org/10.18653/v1/D19-1530} {It{'}s all in the name:
  Mitigating gender bias with name-based counterfactual data substitution}.
\newblock In \emph{Proceedings of the 2019 Conference on Empirical Methods in
  Natural Language Processing and the 9th International Joint Conference on
  Natural Language Processing (EMNLP-IJCNLP)}, pages 5267--5275, Hong Kong,
  China.

\bibitem[{Inoue et~al.(2021)Inoue, Alhafni, Baimukan, Bouamor, and
  Habash}]{inoue-etal-2021-interplay}
Go~Inoue, Bashar Alhafni, Nurpeiis Baimukan, Houda Bouamor, and Nizar Habash.
  2021.
\newblock \href {https://www.aclweb.org/anthology/2021.wanlp-1.10} {The
  interplay of variant, size, and task type in {A}rabic pre-trained language
  models}.
\newblock In \emph{Proceedings of the Sixth Arabic Natural Language Processing
  Workshop}, pages 92--104, Kyiv, Ukraine (Virtual). Association for
  Computational Linguistics.

\bibitem[{Lison and Tiedemann(2016)}]{Lison:2016:opensubtitles2016}
Pierre Lison and J{\"o}rg Tiedemann. 2016.
\newblock Open{S}ubtitles2016: {E}xtracting {L}arge {P}arallel {C}orpora from
  {M}ovie and {TV} {S}ubtitles.
\newblock In \emph{Proceedings of the Language Resources and Evaluation
  Conference (LREC)}, Portoro\v{z}, Slovenia.

\bibitem[{Lu et~al.(2018)Lu, Mardziel, Wu, Amancharla, and
  Datta}]{lu2018gender}
Kaiji Lu, Piotr Mardziel, Fangjing Wu, Preetam Amancharla, and Anupam Datta.
  2018.
\newblock \href {http://arxiv.org/abs/1807.11714} {Gender bias in neural
  natural language processing}.

\bibitem[{Manzini et~al.(2019)Manzini, Yao~Chong, Black, and
  Tsvetkov}]{manzini-etal-2019-black}
Thomas Manzini, Lim Yao~Chong, Alan~W Black, and Yulia Tsvetkov. 2019.
\newblock \href {https://doi.org/10.18653/v1/N19-1062} {Black is to criminal as
  caucasian is to police: Detecting and removing multiclass bias in word
  embeddings}.
\newblock In \emph{Proceedings of the 2019 Conference of the North {A}merican
  Chapter of the Association for Computational Linguistics: Human Language
  Technologies, Volume 1 (Long and Short Papers)}, pages 615--621, Minneapolis,
  Minnesota. Association for Computational Linguistics.

\bibitem[{Nagoudi et~al.(2022)Nagoudi, Elmadany, and
  Abdul-Mageed}]{nagoudi-etal-2022-arat5}
El~Moatez~Billah Nagoudi, AbdelRahim Elmadany, and Muhammad Abdul-Mageed. 2022.
\newblock \href {https://doi.org/10.18653/v1/2022.acl-long.47} {{A}ra{T}5:
  Text-to-text transformers for {A}rabic language generation}.
\newblock In \emph{Proceedings of the 60th Annual Meeting of the Association
  for Computational Linguistics (Volume 1: Long Papers)}, pages 628--647,
  Dublin, Ireland. Association for Computational Linguistics.

\bibitem[{Napoles et~al.(2017)Napoles, Sakaguchi, and
  Tetreault}]{napoles-etal-2017-jfleg}
Courtney Napoles, Keisuke Sakaguchi, and Joel Tetreault. 2017.
\newblock \href {https://aclanthology.org/E17-2037} {{JFLEG}: A fluency corpus
  and benchmark for grammatical error correction}.
\newblock In \emph{Proceedings of the 15th Conference of the {E}uropean Chapter
  of the Association for Computational Linguistics: Volume 2, Short Papers},
  pages 229--234, Valencia, Spain. Association for Computational Linguistics.

\bibitem[{Obeid et~al.(2020)Obeid, Zalmout, Khalifa, Taji, Oudah, Alhafni,
  Inoue, Eryani, Erdmann, and Habash}]{obeid-etal-2020-camel}
Ossama Obeid, Nasser Zalmout, Salam Khalifa, Dima Taji, Mai Oudah, Bashar
  Alhafni, Go~Inoue, Fadhl Eryani, Alexander Erdmann, and Nizar Habash. 2020.
\newblock \href {https://www.aclweb.org/anthology/2020.lrec-1.868} {{CAM}e{L}
  tools: An open source python toolkit for {A}rabic natural language
  processing}.
\newblock In \emph{Proceedings of The 12th Language Resources and Evaluation
  Conference}, pages 7022--7032, Marseille, France. European Language Resources
  Association.

\bibitem[{Papineni et~al.(2002)Papineni, Roukos, Ward, and
  Zhu}]{Papineni:2002:bleu}
Kishore Papineni, Salim Roukos, Todd Ward, and Wei-Jing Zhu. 2002.
\newblock {BLEU}: a {M}ethod for {A}utomatic {E}valuation of {M}achine
  {T}ranslation.
\newblock In \emph{Proceedings of the Conference of the Association for
  Computational Linguistics (ACL)}, pages 311--318, Philadelphia, Pennsylvania,
  USA.

\bibitem[{Post(2018)}]{post-2018-call}
Matt Post. 2018.
\newblock \href {https://doi.org/10.18653/v1/W18-6319} {A call for clarity in
  reporting {BLEU} scores}.
\newblock In \emph{Proceedings of the Third Conference on Machine Translation:
  Research Papers}, pages 186--191, Brussels, Belgium.

\bibitem[{Raffel et~al.(2020)Raffel, Shazeer, Roberts, Lee, Narang, Matena,
  Zhou, Li, and Liu}]{JMLR:v21:20-074}
Colin Raffel, Noam Shazeer, Adam Roberts, Katherine Lee, Sharan Narang, Michael
  Matena, Yanqi Zhou, Wei Li, and Peter~J. Liu. 2020.
\newblock \href {http://jmlr.org/papers/v21/20-074.html} {Exploring the limits
  of transfer learning with a unified text-to-text transformer}.
\newblock \emph{Journal of Machine Learning Research}, 21(140):1--67.

\bibitem[{Rozovskaya et~al.(2015)Rozovskaya, Bouamor, Habash, Zaghouani, Obeid,
  and Mohit}]{rozovskaya-etal-2015-second}
Alla Rozovskaya, Houda Bouamor, Nizar Habash, Wajdi Zaghouani, Ossama Obeid,
  and Behrang Mohit. 2015.
\newblock \href {https://doi.org/10.18653/v1/W15-3204} {The second {QALB}
  shared task on automatic text correction for {A}rabic}.
\newblock In \emph{Proceedings of the Second Workshop on {A}rabic Natural
  Language Processing}, pages 26--35, Beijing, China. Association for
  Computational Linguistics.

\bibitem[{Sennrich et~al.(2016)Sennrich, Haddow, and
  Birch}]{sennrich-etal-2016-controlling}
Rico Sennrich, Barry Haddow, and Alexandra Birch. 2016.
\newblock \href {https://doi.org/10.18653/v1/N16-1005} {Controlling politeness
  in neural machine translation via side constraints}.
\newblock In \emph{Proceedings of the 2016 Conference of the North {A}merican
  Chapter of the Association for Computational Linguistics: Human Language
  Technologies}, pages 35--40, San Diego, California. Association for
  Computational Linguistics.

\bibitem[{Sun et~al.(2019)Sun, Gaut, Tang, Huang, ElSherief, Zhao, Mirza,
  Belding, Chang, and Wang}]{sun-etal-2019-mitigating}
Tony Sun, Andrew Gaut, Shirlyn Tang, Yuxin Huang, Mai ElSherief, Jieyu Zhao,
  Diba Mirza, Elizabeth Belding, Kai-Wei Chang, and William~Yang Wang. 2019.
\newblock \href {https://doi.org/10.18653/v1/P19-1159} {Mitigating gender bias
  in natural language processing: Literature review}.
\newblock In \emph{Proceedings of the 57th Annual Meeting of the Association
  for Computational Linguistics}, pages 1630--1640, Florence, Italy.
  Association for Computational Linguistics.

\bibitem[{Tay et~al.(2021)Tay, Dehghani, Rao, Fedus, Abnar, Chung, Narang,
  Yogatama, Vaswani, and Metzler}]{tay:2021:efficient}
Yi~Tay, Mostafa Dehghani, Jinfeng Rao, William Fedus, Samira Abnar, Hyung~Won
  Chung, Sharan Narang, Dani Yogatama, Ashish Vaswani, and Donald Metzler.
  2021.
\newblock \href {https://doi.org/10.48550/ARXIV.2109.10686} {Scale efficiently:
  Insights from pre-training and fine-tuning transformers}.

\bibitem[{Vaswani et~al.(2017)Vaswani, Shazeer, Parmar, Uszkoreit, Jones,
  Gomez, Kaiser, and Polosukhin}]{Vaswani:2017:attention}
Ashish Vaswani, Noam Shazeer, Niki Parmar, Jakob Uszkoreit, Llion Jones,
  Aidan~N. Gomez, Lukasz Kaiser, and Illia Polosukhin. 2017.
\newblock Attention is all you need.
\newblock \emph{CoRR}, abs/1706.03762.

\bibitem[{Wolf et~al.(2020)Wolf, Debut, Sanh, Chaumond, Delangue, Moi, Cistac,
  Rault, Louf, Funtowicz, Davison, Shleifer, von Platen, Ma, Jernite, Plu, Xu,
  Le~Scao, Gugger, Drame, Lhoest, and Rush}]{wolf-etal-2020-transformers}
Thomas Wolf, Lysandre Debut, Victor Sanh, Julien Chaumond, Clement Delangue,
  Anthony Moi, Pierric Cistac, Tim Rault, Remi Louf, Morgan Funtowicz, Joe
  Davison, Sam Shleifer, Patrick von Platen, Clara Ma, Yacine Jernite, Julien
  Plu, Canwen Xu, Teven Le~Scao, Sylvain Gugger, Mariama Drame, Quentin Lhoest,
  and Alexander Rush. 2020.
\newblock \href {https://doi.org/10.18653/v1/2020.emnlp-demos.6} {Transformers:
  State-of-the-art natural language processing}.
\newblock In \emph{Proceedings of the 2020 Conference on Empirical Methods in
  Natural Language Processing: System Demonstrations}, pages 38--45, Online.
  Association for Computational Linguistics.

\bibitem[{Zhao et~al.(2020)Zhao, Mukherjee, Hosseini, Chang, and
  Hassan~Awadallah}]{zhao2020gender}
Jieyu Zhao, Subhabrata Mukherjee, Saghar Hosseini, Kai-Wei Chang, and Ahmed
  Hassan~Awadallah. 2020.
\newblock \href {https://doi.org/10.18653/v1/2020.acl-main.260} {Gender bias in
  multilingual embeddings and cross-lingual transfer}.
\newblock In \emph{Proceedings of the 58th Annual Meeting of the Association
  for Computational Linguistics}, pages 2896--2907, Online. Association for
  Computational Linguistics.

\bibitem[{Zhao et~al.(2018)Zhao, Zhou, Li, Wang, and
  Chang}]{zhao-etal-2018-learning}
Jieyu Zhao, Yichao Zhou, Zeyu Li, Wei Wang, and Kai-Wei Chang. 2018.
\newblock \href {https://doi.org/10.18653/v1/D18-1521} {Learning gender-neutral
  word embeddings}.
\newblock In \emph{Proceedings of the 2018 Conference on Empirical Methods in
  Natural Language Processing}, pages 4847--4853, Brussels, Belgium.

\bibitem[{Zhu et~al.(2020)Zhu, Xia, Wu, He, Qin, Zhou, Li, and
  Liu}]{zhu-etal-2020-bert-fused}
Jinhua Zhu, Yingce Xia, Lijun Wu, Di~He, Tao Qin, Wengang Zhou, Houqiang Li,
  and Tie-Yan Liu. 2020.
\newblock \href {https://doi.org/10.48550/ARXIV.2002.06823} {Incorporating bert
  into neural machine translation}.

\bibitem[{Zmigrod et~al.(2019)Zmigrod, Mielke, Wallach, and
  Cotterell}]{zmigrod-etal-2019-counterfactual}
Ran Zmigrod, Sabrina~J. Mielke, Hanna Wallach, and Ryan Cotterell. 2019.
\newblock \href {https://doi.org/10.18653/v1/P19-1161} {Counterfactual data
  augmentation for mitigating gender stereotypes in languages with rich
  morphology}.
\newblock In \emph{Proceedings of the 57th Annual Meeting of the Association
  for Computational Linguistics}, pages 1651--1661, Florence, Italy.

\end{thebibliography}
